\documentclass{article}
\usepackage{graph25}

\usepackage{times}
\usepackage{soul}
\usepackage{url}
\usepackage[hidelinks]{hyperref}
\usepackage[utf8]{inputenc}
\usepackage[small]{caption}
\usepackage{graphicx}
\usepackage{amsmath}
\usepackage{amsthm}
\usepackage{booktabs}
\usepackage{algorithm}
\usepackage{algorithmic}
\usepackage[switch]{lineno}
\usepackage{amssymb}
\usepackage{multirow}

\title{Graph Privacy: A Heterogeneous Federated GNN for Trans-Border Financial Data Circulation}

\author{
Zhizhong Tan$^1$
\and
Jiexin Zheng$^{1,2}$\and
Kevin Qi Zhang$^{2,\ast}$ \And
Wenyong Wang$^{1,}$ \thanks{Corresponding author}\\
\affiliations
$^1$Macau University of Science and Technology,Macau,China\\
$^2$Guangdong Institute of Intelligence Science and Technology,Zhuhai,China\\
\emails
zhizhongtan@163.com,
3230002613@student.must.edu.mo,
zhangqi@gdiist.cn,
wywang@must.edu.mo
}

\begin{document}

\maketitle

\begin{abstract}
   The sharing of external data has become a strong demand of financial institutions, but the privacy issue has led to the difficulty of interconnecting different platforms and the low degree of data openness. To effectively solve the privacy problem of financial data in trans-border flow and sharing, to ensure that the data is ‘available but not visible’, to realize the joint portrait of all kinds of heterogeneous data of business organizations in different industries, we propose a Heterogeneous Federated Graph Neural Network (HFGNN) approach. In this method, the distribution of heterogeneous business data of trans-border organizations is taken as subgraphs, and the sharing and circulation process among subgraphs is constructed as a statistically heterogeneous global graph through a central server. Each subgraph learns the corresponding personalized service model through local training to select and update the relevant subset of subgraphs with aggregated parameters, and effectively separates and combines topological and feature information among subgraphs. Finally, our simulation experimental results show that the proposed method has higher accuracy performance and faster convergence speed than existing methods.
\end{abstract}

\section{Introduction}

With the continuous advancement of financial technology's profound empowerment of business, the shared application of external data (such as Internet companies, insurance companies, and other third-party data providers) has become a strong demand for financial institutions. Financial risk control and customer acquisition based on privacy computing have become the most important privacy computing implementation scenario at present \cite{oyewole2024data,farayola2024data}. However, there are three main risks in the current cooperation process between financial institutions and external data sources: first, it involves a large amount of personal user information and is subject to strict regulatory requirements; second, the data assets and trade secrets accumulated by the institution's own business are easy to be leaked; Third, because the data itself can be copied and easily spread, and once shared cannot be traced, the confirmation of data assets is difficult, and commercialization is seriously restricted\cite{christian2024gdpr}.

However, as the current privacy computing technology is still in the process of improvement, the data application in the financial industry still faces various challenges such as the difficulty of interconnection and interoperability of different platforms and the low degree of data openness\cite{zheng2020privacy}. Based on the trans-border trusted data space framework, how to ensure the compliance of data circulation of multiple parties, with the help of privacy computing technology, so that financial institutions and external data partners involved in modeling can achieve the virtual fusion and sample alignment of multi-party data without directly interacting with the original data, and each of them conducts algorithm training locally, and only securely interacts with the intermediate factors of the task, so that the user profiling can be completed without leaving the door of the sensitive data, without going out. At the same time to complete the user's portrait, to achieve financial data trans-border, trans-institutional, trans-sector security sharing and circulation, it has become the current financial data element in the field of the problem to be solved \cite{oyewole2024data}.

\begin{figure}[!tbp]
\centering
\includegraphics[width=1\columnwidth]{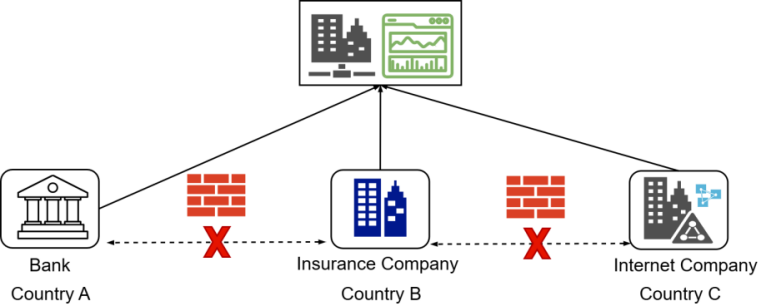}
    \caption{Financial data trans-border sharing scenarios.}
    \label{fig:scenarios}
\end{figure}

At the same time, from the business perspective of trans-border sharing and circulation of financial data, in addition to facing the barriers of many "data islands", we also need to solve the significant problems among "data islands" in the process of collaboration with external data. Heterogeneous issues of business data\cite{yan2024federated}. This is mainly reflected in the variety of data types involved in different financial institutions or industries, the vastly different data formats, and the incompatibility of the data standards and protocols they follow. They are manifested in the fact that there are fewer overlapping users and overlapping characteristics among various participants, as shown in Figure \ref{fig:scenarios}. Therefore, in the face of these complex challenges, our research motivation needs to think about two points: (1) how to use private computing technology to promote the effective trans-border circulation and sharing of financial data;(2) in the process of data circulation and sharing, how to fully mine the potential characteristics of various trans-industry heterogeneous business data to achieve accurate portraits of users.

Given the privacy problems in heterogeneous scenarios with fewer overlapping users and overlapping characteristics among various parties mentioned above, our research motivation is to ensure the compliance of multi-party data sharing under the framework of trans-border trusted data space and break down the barriers of data silos., to solve data privacy and security in trans-border circulation and sharing of financial data, improve the efficiency of data processing and analysis, and promote data cooperation and sharing. Enhance model performance and generalization capabilities and meet compliance and regulatory requirements. To this end, we propose a learning method for Heterogeneous Federal Graph Neural Networks (HFGNN), which aims to solve the privacy issues in the trans-border circulation and sharing of financial data elements in the context of multi-source and heterogeneous service fusion scenarios. In this approach, we combine the respective advantages of GNN and federal learning to address the privacy challenges faced by financial data elements. Experimental results show that our method can effectively solve the personalized privacy issue in the trans-border flow and sharing of multi-source and heterogeneous financial data elements. Our contributions are as follows:

\begin{itemize}
\item We propose a trans-border data circulation and sharing method for heterogeneous federated graph neural networks, which enables model training and data processing without sharing original data, thereby protecting user privacy and data security.   

\item In the process of promoting the trans-border circulation and sharing of financial data elements, we have introduced personalized sub-graph collaboration methods. Based on the diverse characteristics of financial business data, relevant local models within the same branch can be flexibly and targeted collaboratively optimized to meet the needs of different business scenarios. 

\item For multi-source and heterogeneous financial data, we effectively captured the temporal changes and evolution process of dynamic change chart structure data by integrating heterogeneous data between different data sources into a unified feature space and comprehensively analyzing it from the microstructure. The intrinsic logic and value of financial data provide new ideas for financial analysis and decision-making.
\end{itemize}

The subsequent content of this article is organized as follows: In section 2, we discuss in depth the research background of relevant technologies in the circulation and sharing of financial data elements. In section 3 focuses on the key definitions and core concepts involved in our method, laying a theoretical foundation for subsequent content development. In section 4, we elaborate on the overall architecture of our method, the state transition of nodes in the graph and the evolution of the graph structure, the update of weights in model training, and the classification of nodes based on the differentiated needs of financial services. Subsequently, in section 5, we verified the feasibility and effectiveness of the proposed method through simulation experiments. Finally, in section 6, we summarize this paper.

\section{Related Work}

Privacy computing refers to a series of technologies that analyze and calculate data on the premise of ensuring that the data provider does not disclose the original data, ensuring that the data is "available and invisible" during circulation and integration\cite{li2016privacy}. Currently, in terms of research on private computing technology, many researchers mainly focus on exploring and building a private computing financial application ecosystem, with the core of revitalizing data resources and promoting efficient circulation and sharing of data. To this end, this paper sorts out the development process of privacy computing according to the time dimension and roughly divides it into the following four stages. Each stage proposes a new privacy computing solution from a different perspective to try to solve the omissions and flaws in the previous stage\cite{chen2023privacy}.

Phase I: Secure Multiparty Computing. Its development can be traced back to the proposal of Shamir's secret sharing scheme\cite{shamir1979share,sucasas2023secure}, and since then, the field has gradually evolved to form a technical system with secret sharing \cite{yao1982protocols} and obfuscated circuits \cite{yao1986generate} as the core protocols. However, the main problem of secure multi-party computation is that the performance is very different compared to that of plaintext computation. This is mainly due to the practical requirements of the computational and network environments, while its main performance bottleneck lies in the burden due to the vastly increased communication overhead \cite{li2024metamorphic,pillai2024enhancing}.

Phase II: Differential privacy. Unlike cryptography, which relies on theory to prove the difficulty of cracking to ensure security, differential privacy is based on the fuzzification of the data probability distribution caused by adding noise and evaluates its privacy protection effectiveness in a more flexible and changeable dimension \cite{zhao2024scenario}. Because differential privacy does not encrypt the data and does not introduce significant additional communication burdens, its performance is almost close to the efficiency of plaintext computing. Therefore, differential privacy technology is the first to be widely used in various artificial intelligence scenarios, committed to protecting the privacy rights of end-users\cite{wei2023personalized,yang2024local}.

Phase III: Centralized encryption calculation. Unlike the thinking of the previous two phases, the current phase is committed to finding a way to get data out of place while remaining safe. One important technology path is the Trusted Execution Environment (TEE) \cite{islam2023confidential,carreira2024explain,ott2024multitee}. However, the implementation of TEE is highly dependent on hardware support provided by specific vendors. Another technical path is homomorphic encryption \cite{rivest1978data,gentry2009fully,xie2024efficiency,chatel2024veritas}. However, homomorphic encryption requires additional communication and computing costs.

Stage IV: Federal Learning. Different from the above-mentioned technical ideas, federated learning allows each data owner to save the model and data locally, and only transfer protected parameter information between parties to complete the training process. Therefore, private data does not go out of domain under the framework of federal learning, and the exchange of model parameters does not expose the content of raw data and local models. From the perspective of training paradigms, federated learning can be divided into horizontal federated learning, vertical federated learning, and transfer federated learning \cite{brauneck2023federated,lin2024enhancing,gu2023fedpass,sun2021ldp}.

Currently, with the deepening of the research field, in terms of technology, privacy-based encryption computation, and privacy-preserving computation have been widely researched and applied. These techniques are able to achieve the joint analysis and modeling of multi-party data without directly exposing the original data, which provides strong support for critical tasks such as financial risk assessment and fraud detection. In terms of application, privacy computing has been applied on the ground in several scenarios in the financial field. For example, credit risk assessment and anti-fraud areas, etc.

\section{Preliminaries}

\begin{figure}[!tbp]
\centering
\includegraphics[width=1\columnwidth]{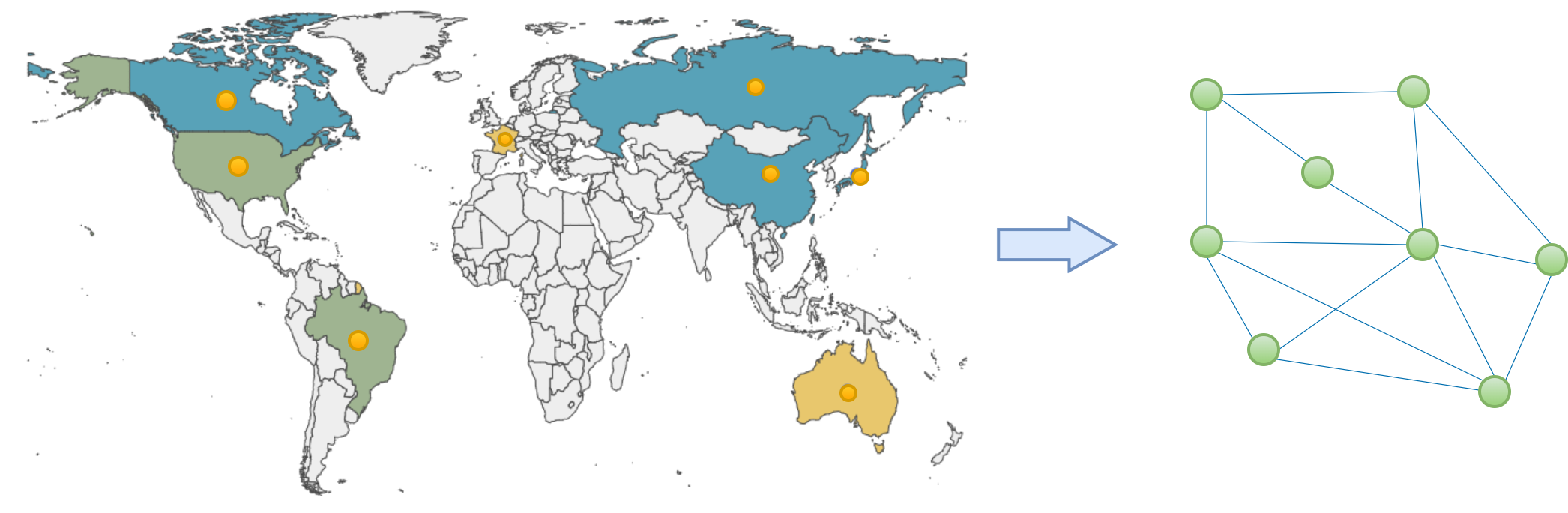}
    \caption{Spatial graph structure: the nodes in the left graph denote the edge servers in different countries in the global trans-border flow of data, which denote the eight countries, such as China, Canada, France, Japan, Russia, United States, Brazil, Australia, etc.; the right graph denotes the graph structure formed by these eight countries in the dynamic process of trans-border data The right figure represents the graph structure of these eight countries in the dynamic process of trans-border data flow, where the nodes represent the edge servers in the countries, the edges represent the transmission paths between the edge servers in the trans-border data flow, and the thickness of the edges represents the size of the transmitted data volume in the data flow or sharing.}
    \label{fig:structure}
\end{figure}

For ease of description, we abstract the edge server relationships between countries or sub-graphs into nodes of the spatial graph structure, and the data flow paths into edges of the spatial graph, as shown in Figure \ref{fig:structure}. In this spatial network structure, each node contains a sub-graph structure, and these sub-graphs are dispersed and deployed on different edge servers. Due to privacy protections or regulatory restrictions in different countries or regions, data on these servers cannot be trained centrally but can be trained collaboratively through federal learning. Information is transferred between the subgraph and the spatial graph through the global server and the boundary server of each subgraph.

\section{Methodology}

\begin{figure*}[!tbp]
\centering
\includegraphics[width=0.75\textwidth,height=0.33\textheight]{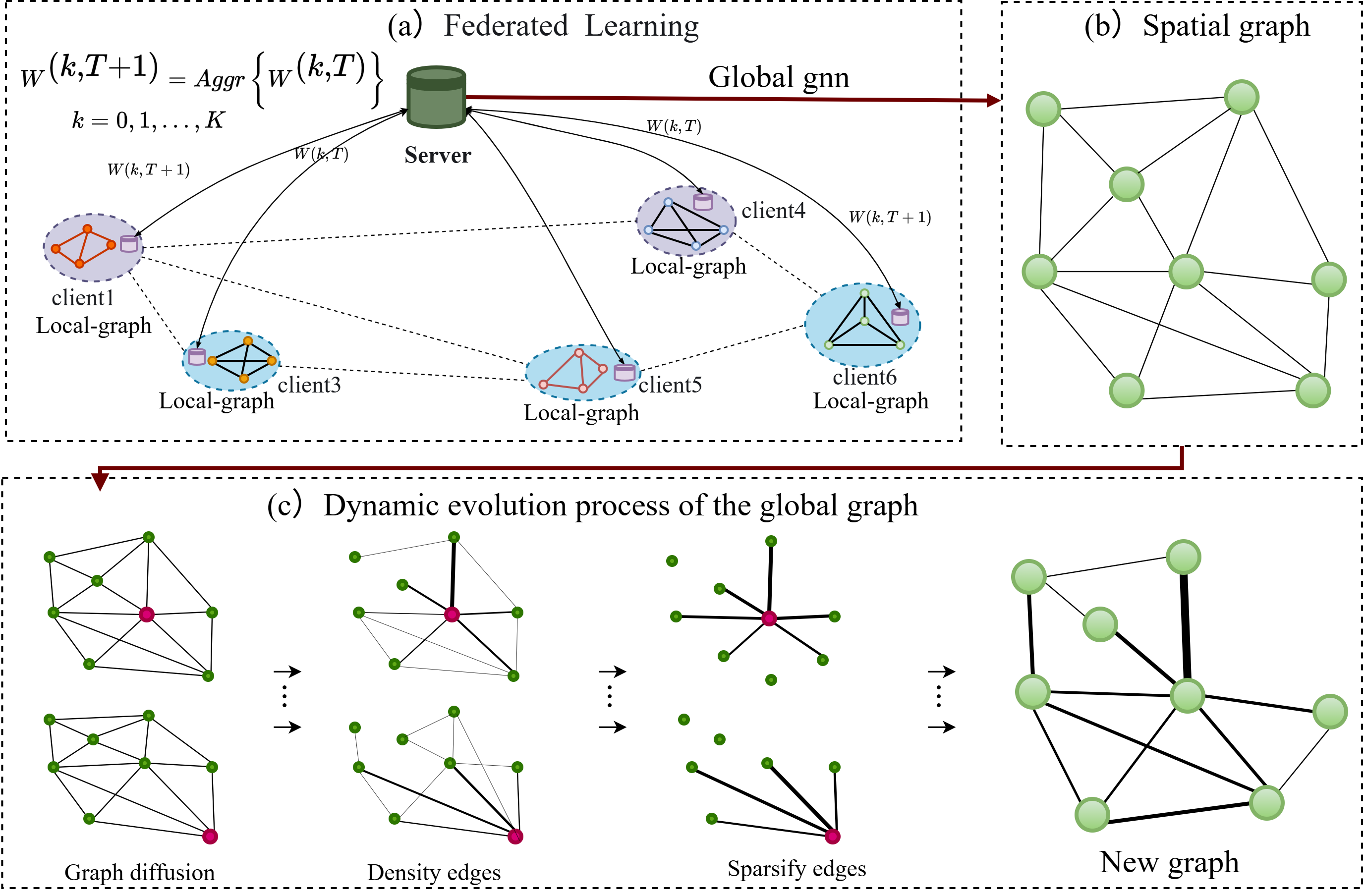}
    \caption{Overview of the proposed model framework.  (a, b) describes the original spatial graph among the eight countries or regions constructed based on their data provider and data demander;  (c) illustrates the process of global graph dynamics evolution, which takes the data provider or demander as input and outputs a refined graph.}
    \label{fig:framework}
\end{figure*}

In the real world, due to the differences in the systems and laws of various countries, how to achieve the free flow and sharing of data under the premise of ensuring data privacy and satisfying different regulatory constraints, and how to use the graph structure information to depict the network structure of the data flow process across domains and borders, is an issue worthy of deep thinking. Since there exists an urgent need for external data sharing applications for financial data elements, the stochastic and subgraph heterogeneity characteristics exhibited in data flow and sharing are also the key areas of concern for us, in addition to the protection of user privacy in trans-border and trans-domain business scenarios of data elements. Because each node in the subgraph is heavily influenced by its relationship with its neighbouring node $N\left (v\right )$, which further affects the dynamic change of the structural features and data statistical features of the subgraph, and any change in the subgraph structure and features will lead to the continuous evolution of the global graph structure over time, it can be seen that the structural states of both the subgraphs and the global graph are constantly evolving. For this reason, we propose a model based on heterogeneous graph federation learning for solving the privacy problem in the trans-border flow and sharing of financial data elements, and the schematic of the framework is shown in Figure \ref{fig:framework}.

According to the design architecture of the model, it is not difficult to know that in the environment of trans-border trusted data space on which the financial data elements depend, the paths of data flow truly reflect the interaction of circulation and sharing between the data demand side and the supply side, which provides us with a good application basis for real-time situational awareness of data. This set of paths can be expressed as $R=\left \{p_{i}\mid i \in \left ( 1,2,\dots,n_{p} \right ) \right \}$, where each path exhibits its unique attributes or characteristics due to business differences between different organizations or subsidiaries.

In this context, we use GNN and federated learning to jointly build and characterize the circulation and sharing process of financial data elements and their privacy calculations. The global GNN represents the overall structure of the dynamic evolution of data flow and sharing and defines it as $G= \left \{V,E,X,W \right \}$ with $n$ nodes, where $V$ represents the set of vertices, $E \subseteq V\times X$ represents the set of edges, $X \in R^{n\times f}$ represents the feature set of nodes, and $W$ represents the training parameters of the model.

Based on the above objectives, each client is responsible for collecting data in its local environment and training local subgraph models that are not visible to other clients. Specifically, we assume that the $k \in \left \{1,2,\dots,K \right \} $ client has its unique private dataset $D^{\left ( k \right ) }:=\left ( G^{\left ( k \right ) },Y^{\left ( k \right ) } \right ) $. In this context, $G^{\left ( k \right ) }:=\left ( V^{\left ( k \right ) },\varepsilon ^{\left ( k \right ) } ,X^{\left ( k \right ) },W^{\left ( k \right ) }\right ) $ denotes the graph in $D^{\left( k \right)}$, $V^{\left( k \right)}$ denotes the set of nodes of the subgraph, $X^{\left ( k \right ) }=\left \{x_{m}^{\left(k\right)}\right \}_{m \in V^{\left(k\right)}} $ and $\varepsilon ^{\left ( k \right ) }=\left \{e_{m,n}^{\left(k\right)}\right \}_{m,n \in V^{\left(k\right)}} $ represent the set of features of the graph data and the corresponding nodes and edges, respectively, $W^{\left(k\right)}$ denotes the client's parameters, and $Y^{\left(k\right)}$ is the set of labels of the graph data. Multiple clients, on the other hand, collaborate through the coordination of a central server, aiming to improve the performance of their respective GNN models. In this process, clients do not need to disclose their local graph datasets, thus ensuring data privacy and security. In this way, the subgraph federated learning system achieves the goal of efficiently utilizing distributed data for model training while protecting privacy.

To achieve our design goals, we will focus on node state transitions, local or global GNN dynamic evolution, model weight updates, and GNN-based node- or edge-level tasks during graph evolution and federated learning involved in the method.

\subsection{Node State Transfer}

Due to the business differences between different institutions or companies, subgraphs have significant heterogeneity characteristics, so each node's state will dynamically evolve with the real-time flow of data every time the model is trained. At any time $t$, the value $Y_{t}$ of the observed variable of each node (such as the data distribution of the node) only depends on the currently hidden state variable $X_{t}$, and is independent of the state variables and observed variables at other times; at the same time, the current state value $X_{t}$ only depends on the state $X_{t-1}$ at the previous time, and is independent of the state. For this reason, we assume that the state variables of nodes are unobservable and hidden, so the subgraph obeys a hidden Markov process in this dynamic evolution process.

\begin{figure}[!tbp]
\centering
\includegraphics[width=1\columnwidth]{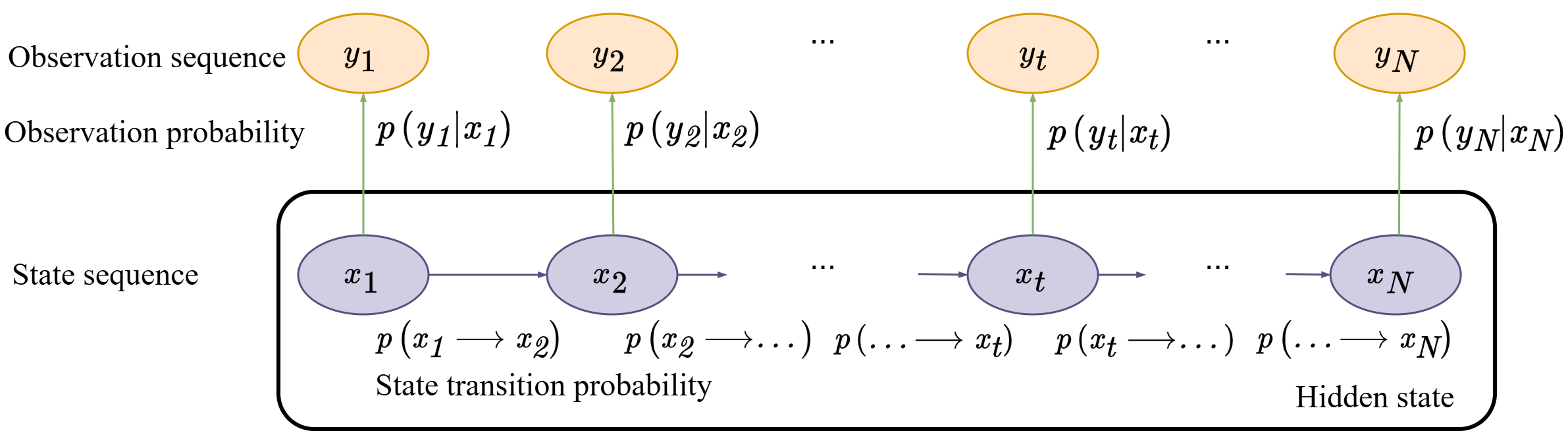}
    \caption{Node state transfer process.}
    \label{fig:transfer}
\end{figure}

As shown in Figure \ref{fig:transfer}, $x\left(t\right)$ is a random variable, a hidden state (invisible) at time $t$, such as $x\left(t\right) \in \left \{x_{1},x_{2},x_{3}\right \}$. And the random variable $y\left(t\right)$ is the value of the observed variable (visible) at time $t$, such as $y\left(t\right) \in \left \{y_{1},y_{2},y_{3},y_{4}\right \}$. The arrows indicate conditional dependencies. Where the random variable $x$ at time $t$, is only conditionally dependent on $x$ at time $t-1$. Similarly, the random variable $y$ at time $t$ is only conditionally dependent on $x$ at time $t$. This property is known as the Markov property.

We assume that the subgraph contains $N$ nodes, representing states; edges represent its transition probability. Based on the hidden Markov model graph structure, the joint probability distribution for all variables is:

\begin{equation}\label{eq:motivation1}
    \begin{split}
    P\left (y_{1},x_{1},y_{2},x_{2},\dots,y_{N},x_{N}\right) \\ =P\left (x_{1}\right)P\left (y_{1}\mid x_{1}\right)\prod_{i=2}^{N}P\left (x_{i}\mid x_{i-1}\right)P\left (y_{i}\mid x_{i}\right)  
    \end{split}
\end{equation}

Where $X=\left (x_{1},x_{2},\dots,x_{N}\right ) $ is a sequence of states and $Y=\left (y_{1},y_{2},\dots,y_{N}\right)$ is a sequence of observations.

\subsection{GNN Dynamic Evolution}

Generally, GNN consists of message propagation and neighborhood aggregation, with each node iteratively collecting information propagated by its neighbors and aggregating it with its information to update its representation. Regarding the dynamic evolution process of the global graph, at time $t$, for client $k$ and layer directory $\ell =0,1,\dots,L-1$, a GNN for a $\ell $-layer can be expressed as:

\begin{equation}\label{eq:motivation1}
    \begin{split}
   msg_{i}^{\left(k,\ell+1\right)}=agg\left(msg_{\theta }^{\left(k,\ell+1\right)}\left(h_{i}^{\left(k,\ell \right)},h_{j}^{\left(k,\ell \right)},\varepsilon _{ij}\right)\mid j\in N_{i}\right)  
    \end{split}
\end{equation}

\begin{equation}\label{eq:motivation1}
    \begin{split}
     h_{i}^{\left(k,\ell+1\right)}=upt_{\phi }^{\left(k,\ell+1\right)}\left (h_{i}^{\left(k,1\right)}, msg_{i}^{\left(k,\ell+1\right)} \right ) 
    \end{split}
\end{equation}

Where, $h_{i}^{\left(k,0\right)}=x_{i}^{\left(k\right)}$ represents the node characteristics of the $k^{th}$ client. $\ell$ is the layer index, which leads to the following: the node characteristics of the $\ell$-layer of the $k^{th}$ client are expressed as $h_{i}^{\left(k,l\right)}$; $agg\left(.\right) $ is an aggregation function that can change according to different GNN variants; $N_{i}$ represents the neighborhood set of node $i$; $agg_{\theta }^{\left(k,\ell+1 \right)}\left (.\right)$ is a message generation function, whose inputs $h_{i}$ are the hidden state of the current node, the hidden state of the neighbor node $h_{j}$, and the edge characteristics $\varepsilon_{ij}$; $upt_{\phi }^{\left(k,\ell+1 \right)}\left (.\right)$ is a state update function used to receive the aggregated feature $msg_{i}^{\left ( k,\ell+1 \right ) }$.

\subsection{Weights Update}

In order to address the data privacy issues of each subgraph, we will construct a strategy for updating parameter weights based on the federated learning framework. We draw on the approach provided in the literature \cite{wang2024rethinking} to enable the model parameters stored on each node to collect and pass information to each other, thus ensuring that knowledge is effectively propagated throughout the graph structure.

Mathematically, given a constructed graph $G_{t}$, for $\forall  k \in K$, we use a weighted sum to update the central node to ensure that the dynamic adjustment of the graph structure is kept in sync with the flow of data.

\begin{equation}\label{eq:motivation1}
    \begin{split}
    \begin{aligned}    
     \widehat{v}_{s,t}^{k} =g\left(\left[v_{s,t}^{1,P-1},\dot,v_{s,t}^{K,P-1}\right] ,\left [\omega_{t}^{k,1},\dot,\omega_{t}^{k,K}\right],P\right) \\     
      = \sum_{i=1}^{K}\omega_{t}^{k,i}\sum_{j=1}^{K}\omega_{t}^{i,j}v_{s,t}^{j,P-2}\\
      = \dots=\underset{k\leftarrow i\leftarrow j \dots \leftarrow u\leftarrow z \left (\# path= P\right ) }{\underbrace{\sum_{i=1}^{K}\omega_{t}^{k,i}\sum_{j=1}^{K}\omega_{t}^{i,j} \dots \sum_{z=1}^{K}\omega_{t}^{u,z} v_{s,t}^{z,0}}}       
    \end{aligned}
    \end{split} 
\end{equation}

Where $g$ is a function used for knowledge dissemination, which utilizes the updated model parameters for information circulation and sharing. $P$ represents the number of repetitions of the dissemination process, which determines the depth and breadth of knowledge dissemination in the graph. $v_{s,t}^{i,P-1}$ is the updated model parameters obtained by node $i$ after $P-1$ rounds of dissemination and aggregation, which incorporates information and wisdom from other nodes. And $\omega_{k,i}$ is the weight between nodes $k$ and $i$, which reflects the importance and influence of these two nodes in the information dissemination process. In particular, when $P=1$, $v_{s,t}^{z,0}=v_{s,t}^{z}$, which indicates the state of model parameters of node $i$ under the initial propagation round. After the above propagation process, we obtain a new set of model collections named $\left\{\widehat{v}_{s,t}^{1},\widehat{v}_{s,t}^{2},\dots,\widehat{v}_{s,t}^{K}\right \} $, which not only contain the information of the clients within the cluster but also incorporate the weighted information of the nodes outside the cluster, enabling the knowledge in the whole graph structure to be comprehensively and accurately propagated and shared.

\section{Experiments}

Based on the method proposed in this paper, we set up corresponding experimental settings closely around the research motivation of this paper. Its main goals focus on model availability and performance. To this end, our experimental design will answer the following key questions: 

\begin{itemize}
\item \textbf{RQ1}: Global situational awareness of trans-border data flows: The goal is to evaluate the dynamic situation of data flows and achieve accurate perception and grasp of the overall situation of data flows by capturing dynamic path changes between data providers and demanders in real-time.  

\item \textbf{RQ2}: Performance Comparison Analysis: The goal is to comprehensively evaluate and verify the performance differences between this method and centralized methods or advanced methods in the current field. On the basis of establishing the usability of this method, we conducted a detailed analysis and comparison of this method with centralized methods and currently advanced methods. The purpose is to visually demonstrate the unique advantages of this method in performance or differences from other methods.
\end{itemize}

\subsection{Experimental Setups}

\paragraph{Datasets.} We evaluated the proposed algorithm by measuring the average test accuracy of two data sets (EMINIST\cite{cohen2017emnist} and CIFAR-10\cite{krizhevsky2009learning}) based on the experimental set-up of literature \cite{jang2022fedclassavg}. To conduct experiments evenly and fairly, in each setting, the client has a corresponding dataset and randomly divides it into three parts: 80\% for training, 10\% for verification, and 10\% for testing. In this experiment, we adjusted the local model parameters for each client based on the accuracy and efficiency of the model on the verification set. Through training and analysis of the above data, the privacy and usability of the model under the privacy protection mechanism are evaluated.

\paragraph{Baseline.} In our experiments, we chose the following baselines for comparison: (1) local each client trains their model locally, and (2) globally servers are trained using the complete graph. Meanwhile, to fairly evaluate the proposed HFGNN, the following classical federated learning model baselines were also selected: fedAvg \cite{mcmahan2017communication}, fedProx \cite{li2020federated}, and fed-pub \cite{baek2023personalized}. Among them, to further validate the robustness and scalability of HFGNN, we set the number of clients ranging from 5, 10, and 30.

\paragraph{Implementation Details.} In our experiments, we used ResNet-18\cite{he2016deep} as the basic model for three image datasets. All baselines and HFGNNs use the same network to maintain fairness in comparisons. For all networks, we use Adam\cite{kingma2014adam} as the optimizer of choice, with a momentum of 0.0001 and a weight decay of $5e-4$. The number of communication rounds for all data sets is 100, and the number of local training times is 5. In addition to the above information, the parameter settings for training the local model, such as learning rate, training sample size, optimization algorithm used, etc., are shown in Table \ref{tab:booktabs}.

\begin{table}
    \centering
    \begin{tabular}{ccccc}
        \toprule
        Dataset  &$\#$epochs &Batch size &$\rho$ &Learning rate \\
        \midrule
        EMINIST     &1    &32    &0.1   &0.0004     \\
        CIFAR-10    &1    &32    &0.1   &0.0001   \\
        \bottomrule
    \end{tabular}
    \caption{Hyperparameters used for local client update.}
    \label{tab:booktabs}
\end{table}

\subsection{Experimental Results}

\begin{figure}[!tbp]
\centering
\includegraphics[width=1\columnwidth]{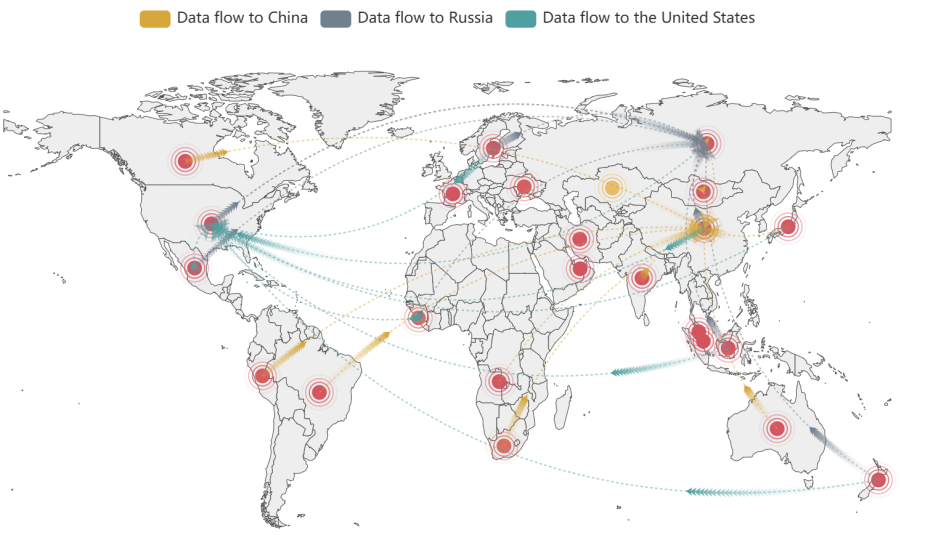}
    \caption{Dynamic evolutionary dynamics of global maps.}
    \label{fig:maps}
\end{figure}

\paragraph{Global diagram structure.} To further explore the effectiveness of the proposed HFGNN, we designed a case study to demonstrate the dynamic evolution of the global graph. We are interested in testing whether our model can capture and rebuild hidden relationships between local clients through iterative updates in the HFGNN framework. Here, we first randomly divide each data set into multiple parts and allocate them to different participants in the federated learning network, so that each client holds a part of the data set. To simulate the distribution structure of global data among different regions or institutions and depict the flow process of data among participants (such as data uploading, downloading, aggregation, etc.), we record key information about data flow during the simulation process, such as flow time, data volume, participants, etc. Thus, the interactive information of these data flows is used to simulate the dynamic evolution process of the global structure diagram, as shown in Figure \ref{fig:maps}.

\paragraph{Performance comparison.}

\textbf{(1) Compared with centralized training.} To compare the effects of federal training and centralized training, we set the configuration files according to the following parameters: 

\textbf{(a) Federal training configuration.} Total number of client devices: 10 (num\_models=10); number of devices participating in each round of training: randomly selected 5 of them (client\_k=5); number of local training iterations: each device iterates 5 times each time it participates in training (1ocal\_epochs=5); number of global iterations: 20 rounds of global iterations are conducted throughout the training process (global\_epochs=20). 

The purpose of this configuration is to comprehensively evaluate the training effect achieved by decentralizing computing resources under the federated learning framework.

\textbf{(b) Centralised training configuration.} There is no need to write centralized training code separately. You only need to adjust the federal learning configuration to achieve the same effect as centralized training. Specific adjustments are as follows: 

Set the total number of client devices (num\_models) and the number of devices participating in training per round (client\_k) to 1 to simulate a federal training scenario in which only one device participates, thus achieving the equivalent of centralized training; The number of local training iterations is set to 1 (local\_epochs=1) to conform to the characteristics of centralized training; The remaining parameter configurations remain consistent with the federated learning training to ensure the fairness and accuracy of the comparative experiment. 

\begin{figure}[!tbp]
\centering
\includegraphics[width=1\columnwidth]{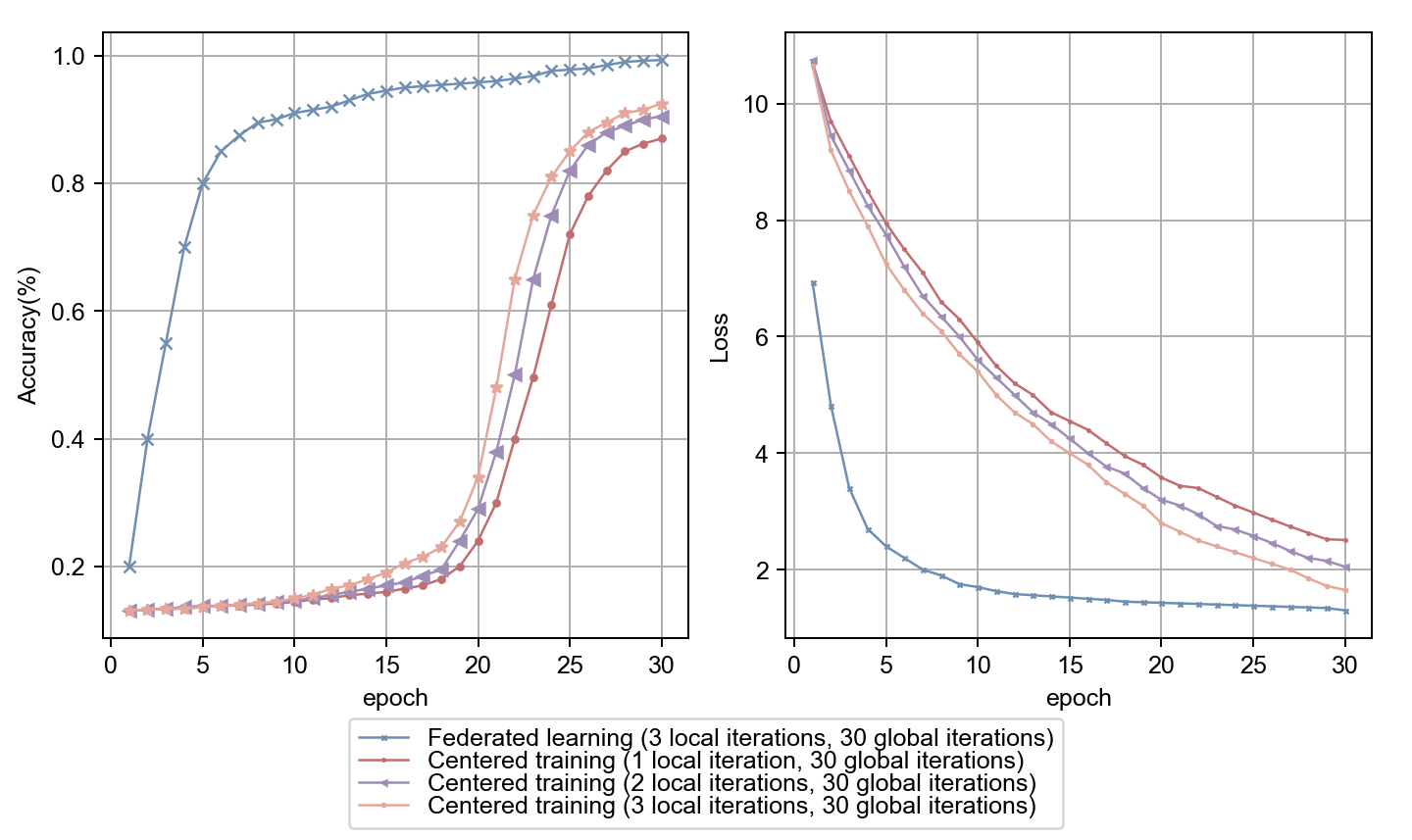}
    \caption{Comparison of the results of the two training methods after 30 rounds of iterations, with the accuracy comparison on the left and the loss function value comparison on the right.}
    \label{fig:Comparison}
\end{figure}

Figure \ref{fig:Comparison} visually shows the performance comparison of two different training methods in the CIFAR-10 image classification task. It can be observed from the figure that the training effectiveness of federal learning is almost equal to that of centralized training, showing a high degree of similarity.

\begin{figure}[!tbp]
\centering
\includegraphics[width=1\columnwidth]{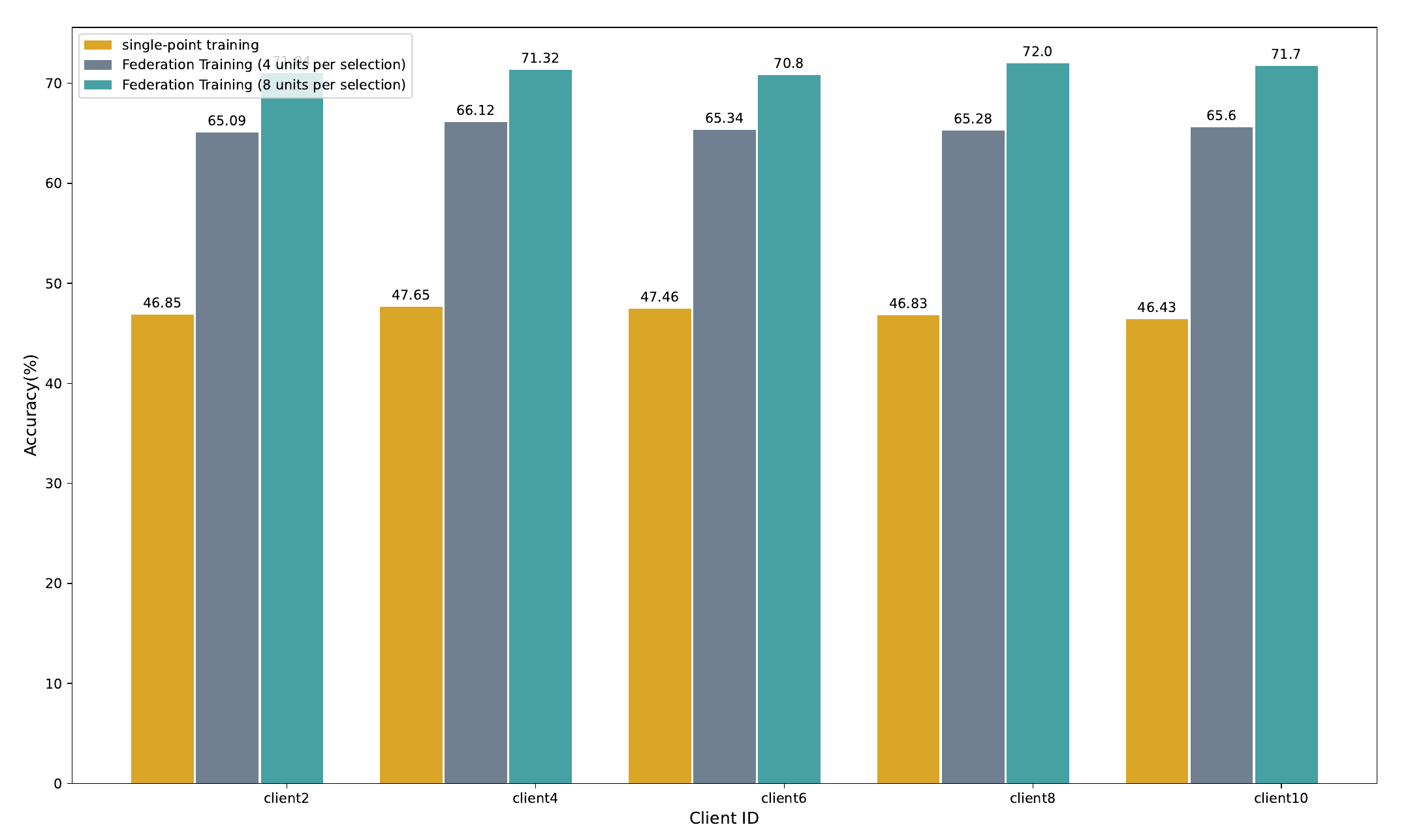}
    \caption{Comparison of the performance of federally trained models with single-point trained models in the inference phase.}
    \label{fig:single-point}
\end{figure}

\textbf{(2) Comparison with single point training model.} Figure \ref{fig:single-point} illustrates the comparison of model performance in the inference phase. A single-site training model is a model that is trained locally and iteratively on a specific single client $C$, using its unique local dataset $D$. For comparison, we randomly selected five different clients, each of which was trained independently on a single point. In the federated training scenario, we separately set different $k$ values to represent each local iteration of training by selecting the number of participating clients from all clients. In this experiment, we especially set two scenarios of $k=4$ and $k=8$ for comparison to observe the effect of different numbers of participating clients on the effect of federation training.

In the presentation of Figure \ref{fig:single-point}, we can see that the effect of training the model on a single client (orange bars) is significantly lower than the effect of training the model on a federation (light grey and light blue bars). This phenomenon strongly suggests that it is difficult to adequately capture the global distributional features of the data when training with only a single client's data, which leads to a relatively poor generalization ability of the model. In addition, it is worth noting that the number of clients involved in each round of federated training (i.e., k-value) has a certain impact on the model performance. Specifically, as the k-value increases, it means that the number of participating clients in each round of training increases, which usually leads to performance improvement. However, at the same time, the completion time of each training round will be extended accordingly. Therefore, in practice, the relationship between the number of participating clients and the training efficiency needs to be weighed to find the optimal balance.

\textbf{(3) Performance comparison with other models.} We do this by varying the number of subgraphs into which each graph is divided, i.e., 5, 10, and 30 subgraphs, each owned by a single client. Table \ref{tab:ablation_study} gives the performance results of different algorithms under three datasets. We can see that HFGNN outperforms all other baselines under different datasets, which confirms the excellent ability of HFGNN to handle heterogeneous scenarios with multiple sources. The reason behind this is that HFGNN can share generic knowledge across similar clients and capture personalized domain-specific knowledge across different clients. In contrast, the traditional federated learning algorithm, FedAvg, performs slightly better than the Local method but still has room for improvement because it does not sufficiently take into account the heterogeneity of data. The FedProx algorithm, on the other hand, improves the shortcomings of FedAvg to some extent. The personalized FL algorithm Fed-pub shows better performance than normal aggregation by using a generic solution to the non-iid problem.

\begin{table}[!tbp] 
  \centering
	  \resizebox{\columnwidth}{!}{      
        \begin{tabular}{lccccccccccc}
        \toprule
        \rule{0pt}{12pt}
        \multirow{2}{*}{Methods}&
        \multicolumn{3}{c}{EMINIST}&\multicolumn{3}{c}{CIFAR-10}\cr 
        \cmidrule(lr){2-4} \cmidrule(lr){5-7} 
        &M=5 &M=10 &M=30 &M=5 &M=10 &M=30 \cr
        \midrule
        Global		& 	&76.91±1.02	&   	& 	&88.38±0.33	& \cr
        Local 		&73.85±1.20	&48.91±2.34	&64.54±0.42	&83.81±0.69	&59.19±1.31	&80.72±0.16\cr
		FedAvg		&76.37±0.43	&75.92±0.21	&66.15±0.64	&85.29±0.83	&84.57±0.29	&82.05±0.12\cr
		FedProx		&77.15±0.45	&76.87±0.80	&66.11±0.75	&85.21±0.24	&84.98±0.65	&82.13±0.13\cr
		Fed-pub 	&93.22±0.07 &91.60±0.08	&83.00±0.06	&80.50±0.15 &81.20±0.13	&82.00±0.10\cr
		Ours		&94.00±0.07	&93.60±0.06	&85.20±0.05	&89.20±0.14	&88.90±0.12	&84.50±0.08\cr
        \bottomrule
        \end{tabular}}
    \caption{Comparison with the state-of-the-art methods on EMINIST and CIFAR-10 datasets.} 
    \label{tab:ablation_study}
\end{table}

In addition, our HFGNN method shows significant advantages in dealing with the fusion problem of multi-source heterogeneous client information. When the number of subgraphs (i.e. clients) further increases, the performance of most methods will be significantly affected and a significant decline will occur due to the lack of information between clients. However, as the number of subgraphs increases, the GNHFN method can still maintain excellent performance and a high degree of robustness. This is due to the unique ability of the HFGNN method to accurately identify potentially related subgraphs and effectively capture important information missing between clients based on this.

\section{Conclusion}

In this paper, we discuss the privacy protection issues in the sharing and circulation of multi-source and heterogeneous business data in the trans-border data space from the business level of financial data elements and recognize two main challenges, including trans-border data privacy issues and Effective integration of heterogeneous business data in the collaborative process of sharing and circulation. Therefore, to overcome these challenges, we proposed HFGNN as a graph federation learning framework that can help each client improve the performance of personalized local models. The key idea of HFGNN is that we can effectively separate and combine the topological information and feature information of multiple heterogeneous subgraphs. We first use the topological feature information of the customer terminal diagram as a joint representation to obtain better representation capabilities. Then on the server side, we separate the aggregation process of topological feature parameters separately, to allow each client to learn personalized local models. Finally, our simulation results show that HFGNN has higher accuracy performance and faster convergence speed than existing methods.

\bibliographystyle{named}
\bibliography{GraphPrivacy}

\end{document}